%% file: main.tex
\documentclass[10pt,twocolumn,letterpaper]{article}

\usepackage[final]{iccv}      

\usepackage{alltt}
\usepackage{graphicx}
 \usepackage{multirow} 
\usepackage{microtype}


\definecolor{iccvblue}{rgb}{0.21,0.49,0.74}
\usepackage[pagebackref,breaklinks,colorlinks,allcolors=iccvblue]{hyperref}

\def\paperID{6955} 
\def\confName{ICCV}
\def\confYear{2025}

\newcommand{\ourmodel}{DiTaiListener\xspace}

\title{\ourmodel: Controllable High Fidelity Listener Video Generation with Diffusion}

\author{
\href{https://cv.maxi.su/}{\textcolor{black}{Maksim Siniukov\textsuperscript{*}}} \quad
\href{https://boese0601.github.io/}{\textcolor{black}{Di Chang\textsuperscript{*}}} \quad
\href{https://scholar.google.com/citations?hl=en\&user=HuuQRj4AAAAJ}{\textcolor{black}{Minh Tran}} \\
\href{https://www.linkedin.com/in/hongkun-kevin-gong-7296a8262/}{\textcolor{black}{Hongkun Gong}} \quad
\href{https://ashutoshchaubey.in/}{\textcolor{black}{Ashutosh Chaubey}} \quad
\href{https://people.ict.usc.edu/~soleymani/}{\textcolor{black}{Mohammad Soleymani}} \\
University of Southern California \\
Los Angeles, USA\\
\href{http://havent-invented.github.io/DiTaiListener}{havent-invented.github.io/DiTaiListener} \\
{\tt\small siniukov@usc.edu}
}

\begin{document}
\twocolumn[{%
\renewcommand\twocolumn[1][]{#1}%
\maketitle
    \captionsetup{type=figure}
    \vspace{-9mm}
    \includegraphics[width=\textwidth]{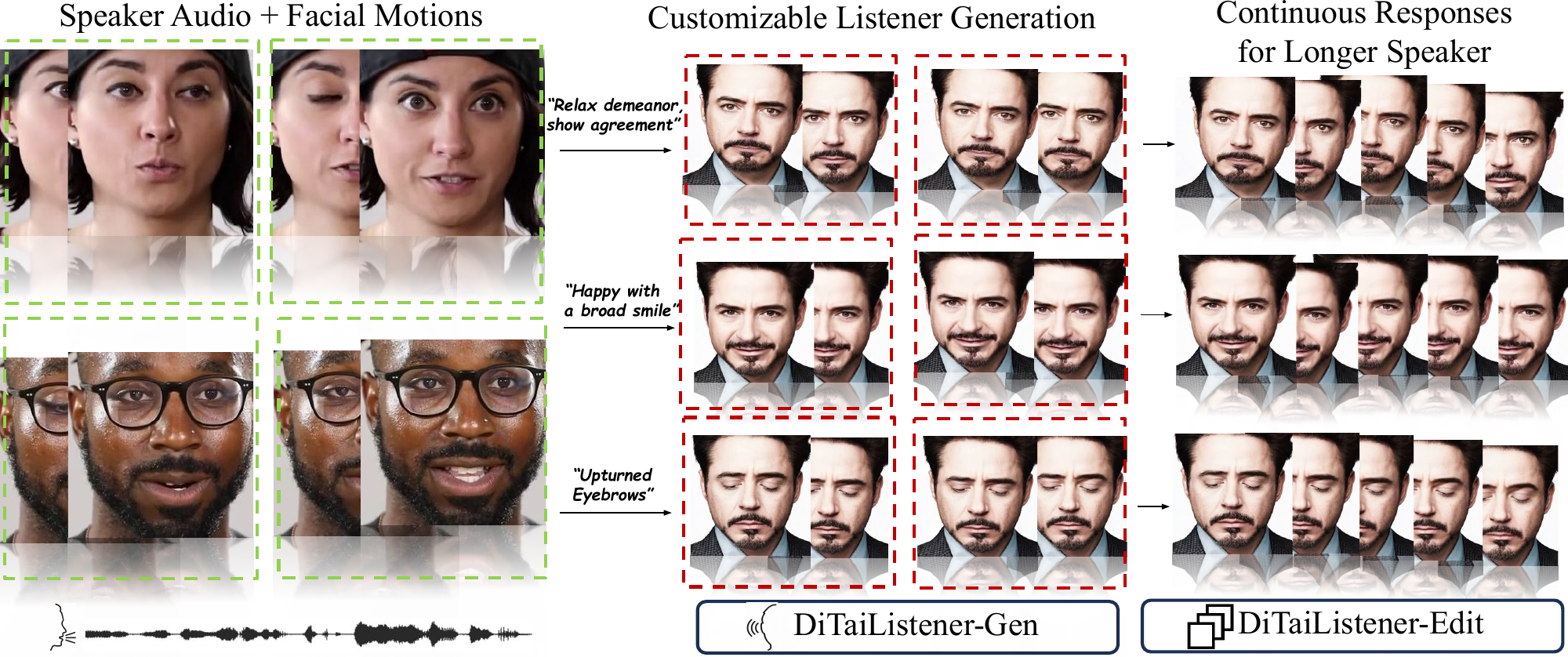} 
    \vspace{-5mm}
    \hfill\caption{
    We introduce \ourmodel, a DiT-based listener generation model that synthesizes high-fidelity human portrait videos of listener behaviors from speaker audio and facial motion inputs in an end-to-end manner. \textit{\ourmodel-Gen} generates customizable listener responses in short segments, while \textit{\ourmodel-Edit} ensures seamless transitions between segments, producing continuous and natural listener behaviors. Together, \ourmodel supports user-friendly customizable listener behavior generation for variable speaker inputs.
    }
    \label{fig:teaser}
    \hfill \vspace{0mm}
}]
\def\thefootnote{*}\footnotetext{Equally contributed as first authors}

\input{sec/0_abstract}    
\input{sec/1_intro}

\input{sec/2_related}

\input{sec/3_method}
\input{sec/4_exp}

\input{sec/5_conclusion}
\input{sec/ackn}

\clearpage
\newpage
{
    \small
    \bibliographystyle{ieeenat_fullname}
    \bibliography{main}
}

\input{sec/X_suppl}

\end{document}

%% file: sec/0_abstract.tex
\begin{abstract}
Generating naturalistic and nuanced listener motions for extended interactions remains an open problem.
Existing methods often rely on low-dimensional motion codes for facial behavior generation followed by photorealistic rendering, limiting both visual fidelity and expressive richness.
To address these challenges, we introduce \ourmodel, powered by a video diffusion model with multimodal conditions. Our approach first generates short segments of listener responses conditioned on the speaker's speech and facial motions with \ourmodel-Gen. It then refines the transitional frames via \ourmodel-Edit for a seamless transition. 
Specifically, \ourmodel-Gen adapts a Diffusion Transformer (DiT) for the task of listener head portrait generation by introducing a Causal Temporal Multimodal Adapter (CTM-Adapter) to process speakers' auditory and visual cues. CTM-Adapter integrates speakers' input in a causal manner into the video generation process to ensure temporally coherent listener responses. 
For long-form video generation, we introduce \ourmodel-Edit, a transition refinement video-to-video diffusion model. The model fuses video segments into smooth and continuous videos, ensuring temporal consistency in facial expressions and image quality when merging short video segments produced by \ourmodel-Gen.
Quantitatively, \ourmodel achieves the state-of-the-art performance on benchmark datasets in both photorealism (+73.8\% in FID on RealTalk) and motion representation (+6.1\% in FD metric on VICO) spaces.
User studies confirm the superior performance of \ourmodel, with the model being the clear preference in terms of feedback, diversity, and smoothness, outperforming competitors by a significant margin.
\end{abstract} 

%% file: sec/1_intro.tex
\section{Introduction}
Human behavior synthesis is an essential building block for socially intelligent systems, with broad applications from health to entertainment. Hence, audio-driven head motion generation has received considerable attention due to its potential for generating engaging content \cite{PC-AVS,ng2022learning,ng2023can,song2023emotional,tian2024emo,xu2024vasa,liu2024customlistener,wang2023agentavatar,tran2024dim,zhou2023interactive,guan2023stylesync}. A grand majority of work has focused on speaker head generation \cite{xu2024vasa,zhang2023sadtalker,tian2024emo,chen2024echomimic,xu2024hallo,zhang2023dream,liu2024anitalker,wang2024v,RAD-NeRF,PC-AVS,Wav2Lip,MakeItTalk,zhang2024personatalk} synthesizing speakers’ facial motions based on speech input, producing vivid lip synchronization yet neglecting listeners’ nonverbal feedback or conversational context. Coordinated and contextual listener behaviors are essential to building a sense of rapport, a crucial element of human interpersonal communication with an impact on interaction quality and trust \cite{gratch2006virtual}. Hence, Listening-head generation \cite{liu2024customlistener,ng2022learning,ng2023can,song2023emotional,liu2023mfr,zhao2022semantic,zhou2022responsive,react2024,tran2024dim,zhu2024infp} focuses on coordinated listeners' responses to speaker behaviors, yet typically restricts them to limited reactive facial movements in a limited temporal context, i.e., responding to immediate speaker behaviors with limited motions. While some recent efforts \cite{tran2024dim,zhou2023interactive,wang2023agentavatar,liu2024customlistener,zhu2024infp} have begun exploring the dyadic context in generating reactions, most continue to rely on motion tokenization rather than generating high-fidelity portraits with nuanced expressions and facial details. 

We propose \ourmodel, an end-to-end video diffusion-based listener head generation framework. Through extensive experiments, we study the potential of video generation models in learning nuanced and coordinated facial motions that are well-timed, detailed, and controllable. While prior methods generate listener behaviors in a compact latent space, e.g., 3D Morphable Models (3DMMs) or other motion codes, and then render the face, inevitably losing detail and expressiveness, we directly predict the listener’s behavior in a higher-fidelity video space through diffusion, offering more varied and fine-grained outputs. Large Video Diffusion models capture emergent, data-driven behaviors, producing more dynamic and lifelike motions, including natural blinking patterns and subtle facial expressions. Our method also supports flexible controls from users; that is, once emotion guidance or response context is specified through a text prompt, the model can generate listener reactions accordingly. 

Reactivity in human interactions is not strictly synchronous; instead, listeners react to events in the temporal context with uncertain lag. For example, smile mimicry occurs almost immediately, whereas a head nod is usually delayed, aligning more with speaker pauses as a non-rigid response. To model this, we use a temporal attention mechanism that learns to adaptively time the reactions. This enables \ourmodel to learn better response timing. Our design incorporates a causal temporal adapter, which aligns preceding speaker cues to predicted listener motions, improving temporal coherence, in addition to long-term speaker audiovisual behavior encoders that gather context from up to 10 seconds of speaker video. This architecture allows for the generation of extended video sequences, including those lasting minutes or more, via \ourmodel-Edit, a temporal frame-smoothing approach. Experimental evaluations show that \ourmodel achieves state-of-the-art results in listener behavior synthesis according to both quantitative measures and human evaluations.

Our main contributions can be summarized as follows.
\begin{itemize}
    \item We propose an end-to-end listener generation model that can synthesize listener motions in pixel space and demonstrate the potential of video generation methods for socially intelligent, coordinated, and controlled face and head gestures.
    
    \item We introduce a causal temporal multimodal adapter (CTM-Adapter) that aligns listener motions with preceding audiovisual speaker behaviors. This improves the temporal consistency between speaker cues and the listener’s responses.    
    
    \item Our model provides free-form text control to guide the listener's behaviors. Given emotional guidance and/or dialog context, the model can produce listener reactions according to the provided guidance.
       
    \item We develop a long-sequence video generation approach through frame smoothing with \ourmodel-Edit for variable-length listener videos.
\end{itemize}
\begin{figure*}[ht]
    \centering
       \includegraphics[width=\linewidth, trim=20 0 20 0, clip]{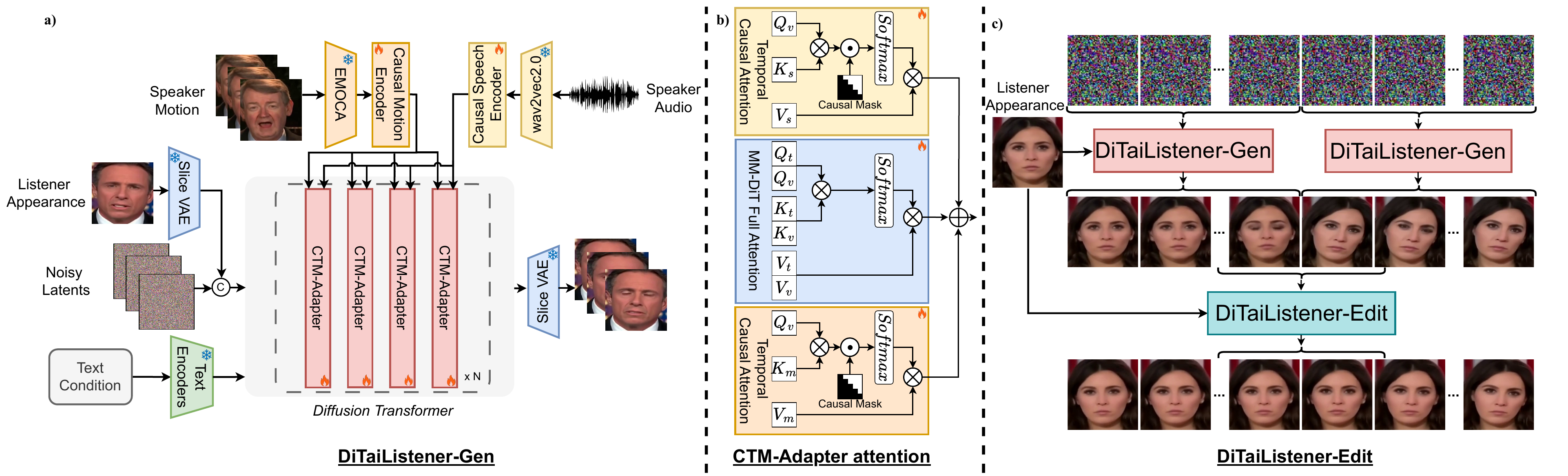}
    \caption{\textbf{Overview of \ourmodel.} a) Given the listener's appearance (reference frame), speaker's motion, encoded via EMOCA 3DMM coefficients, speech (Wav2Vec2) and an input text control, \ourmodel learns to generate listener face and head motions in pixel space through a video diffusion model powered by a modified DiT. b) We introduce a Causal Temporal Multimodal Adapter for seamless integration of multimodal speaker input in a temporally causal manner. c) Our long video generation pipeline consists of two video generation models. \ourmodel-Gen generates video blocks that are fused by the \ourmodel-Edit model that facilitates the smooth transition between two blocks, improving smoothness and reducing computational cost compared to existing long video generation strategies, e.g., prompt traveling and teacher forcing.}
    \label{fig:pipeline}
\end{figure*}

%% file: sec/2_related.tex

\section{Related Work}\label{sec:related}
\subsection{Diffusion-based Portrait Generation}
Recent advances in Diffusion Probabilistic Models (DMs) \cite{ho2020denoising,song2020denoising,song2020score} have exhibited extraordinary performance in a variety of generative tasks, such as image synthesis \cite{saharia2022photorealistic}, video generation~\cite{blattmann2023stable}, and multi-view rendering~\cite{liu2023zero1to3, liu2023one2345, gu2023diffportrait3d}. Latent diffusion models (LDMs)~\cite{rombach2021highresolution} build on these successes by conducting the diffusion process within a lower-dimensional latent space, thereby substantially reducing computational overhead.
In the realm of portrait animation, pre-trained diffusion models~\cite{saharia2022photorealistic,rombach2021highresolution} have served as a strong foundation for image-to-video (I2V) generation. A number of previous works~\cite{cao2023masactrl,lin2023consistent123,Zhangreference} have highlighted the efficacy of infusing reference image features into the self-attention layers of LDM UNets, enabling both image editing and video generation while preserving appearance consistency. Additionally, ControlNet~\cite{zhang2023adding} expands LDM-based generation by incorporating structural inputs—such as landmarks, segmentations, and dense poses—into the model. Leveraging these strategies, recent research~\cite{hu2023animateanyone,xu2023magicanimate,chang2024magicpose} has demonstrated state-of-the-art full-body animation by integrating appearance features, motion control modules, and temporal attention mechanisms~\cite{guo2023animatediff,guo2023sparsectrl} within ReferenceNet architectures.
Subsequent works~\cite{xu2024hallo,cui2024hallo2,wei2024aniportraitaudiodrivensynthesisphotorealistic,ma2024follow} have extended portrait animation to accommodate a range of facial motion representations and speech signals. Nevertheless, these methods currently remain confined to one-way portrait animation and cannot yet produce convincing two-way conversational interactions between speakers and listeners.

\subsection{Dyadic Behavior Generation} 
Unlike human portrait generation, e.g., talking heads, listener behavior generation in dyadic settings focuses on generating a listener’s response to the speaker's verbal and nonverbal behaviors. Early work relied on rule-based methods and simple machine learning models to generate discrete behaviors for virtual humans, e.g., a head nod or smile \cite{rapportAgent}, and heavily relied on the speaker's speech. Huang et al.~\cite{huang2011virtual} extended the rule-based listener response generation with a conditional random field (CRF) model to generate different kinds of head nods. Bohus and Horvitz~\cite{bohus2010facilitating} studied the effect of facial nonverbal behavior generation for facilitating multiparty turn-taking. Greenwood et al.~\cite{greenwood2017predicting} studied speech-driven synchronized agent motion generation in dyadic settings. Ahuja et al.~\cite{ahuja2019react} focus on producing non-verbal full-body motions in dyadic interactions by integrating monadic (within-person) and dyadic (between-person) behaviors to predict socially appropriate body poses. Other techniques aim to generate realistic listener facial movements based on language \cite{chu2018face} or speech signals~\cite{jonell2019learning,jonell2020let}.
For example, Song et al.~\cite{song2023emotional} proposed Emotional Listener Portrait (ELP) that learns to generate nonverbal listener motions that can be modulated by emotions. Geng et al. ~\cite{geng2023affective} introduced the RealTalk database and proposed a method to retrieve plausible listener expressions by leveraging a large language model to condition motions on listeners' goals, personalities or backgrounds.
Ng et al.~\cite{ng2022learning} proposed Learning2Listen (L2L) that relies on learning a dictionary of listener motions through VQ-VAE \cite{van2017neural} and generating listener motions based on the speaker's motion and speech in an autoregressive manner. Tran et al.~\cite{tran2024dim} proposed Dyadic-Interaction-Modeling (DIM) that employs a similar approach to predict discrete listener head motions but leverages self-supervised pre-training with a large dataset to improve representations and uses PIRenderer~\cite{ren2021pirenderer} for photorealistic synthesis. More recently, CustomListener~\cite{liu2024customlistener} proposed a text-guided generation strategy that encodes user-defined textual attributes into portrait tokens containing time-dependent cues for speaker-listener coordination and then feeds these tokens into diffusion-based motion priors for controllable video generation. Zhu et al.~\cite{zhu2024infp} introduce INFP that takes a two-stage audio-driven approach, learning a low-dimensional motion latent space and mapping dual-track audio to these latent codes via a diffusion-based transformer. Although these diffusion-based methods enable responsive and natural dyadic interactions, they predict the motion latent codes that need to be rendered through a warping mechanism, posing challenges for directly generating highly diverse and detailed results in video space. Unlike previous work, \ourmodel learns to generate listener motions directly in the pixel space, providing a more realistic and detailed appearance and motion.

%% file: sec/3_method.tex
\section{Method}



We provide an overview of \ourmodel in Figure \ref{fig:pipeline}. Given the speaker’s audio and facial motions and the listener's reference frame (identity), our model synthesizes high-fidelity, photorealistic images of the listener’s expected facial response. Unlike most prior work that first predicts 3DMM features for the listener and then applies photorealistic rendering, \ourmodel introduces an end-to-end solution that directly synthesizes high-quality listener facial images via a standard diffusion denoising process with a novel diffusion transformer architecture built upon DiT. Our method not only enhances realism but also enables text-based control for customizing listener responses, offering greater flexibility and expressiveness in facial reaction generation. At the core of our approach is the Causal Temporal Multimodal Adapter (CTM-Adapter), a module integrated into the DiT architecture to process multimodal inputs—speaker audio and facial motion features—in a temporally causal manner while enabling text-based control for customized listener responses. We further enhance \ourmodel by introducing seamless transitions between independently synthesized video segments, allowing for the generation of long (unbounded), coherent listener response videos, a limitation of prior video generation methods. In the following sections, we provide details for each component of \ourmodel. 

\subsection{Visual and behavioral Descriptors}
\noindent \textbf{Speaker Feature Representations.} For speaker motions, we follow prior work \cite{learning2listen, tran2024dim, ng2023can} and utilize EMOCA \cite{emoca} to extract facial representations at a 12 FPS frame rate. For speaker audio, we employ a pre-trained Wav2Vec2-base \cite{baevski2020wav2vec} model to capture rich speech representations. Specifically, we leverage the intermediate features from all 12 layers of Wav2Vec2 to preserve detailed acoustic information. To ensure temporal alignment, we resample the extracted speech representations to match the frame rate of the speaker motion features. The features are passed to transformer-based causal motion encoders and causal speech encoders to extract relative information. We denote the resulting speaker speech representation as $X_s$, and the speaker facial motion features as $X_m$. \\
\noindent \textbf{Listener identity.} We utilize features extracted from Slice-VAE \cite{xu2024easyanimate} to encode the reference listener image into the diffusion latent space. This approach is commonly used for spatial conditioning in image and video generation through depth maps and edge maps \cite{xu2024easyanimatehighperformancelongvideo}. We extend this approach to provide appearance guidance. We denote the extracted identity representation as $X_i$. \\
\noindent \textbf{Customized text prompts.} \ourmodel optionally accepts text prompts to enable controllable listener behavior generation. When a text prompt is provided, we encode it using the T5 \cite{2020t5} and BERT~\cite{devlin2019bert} text encoders, allowing for flexible customization of listener responses. We denote the extracted text representation as $X_t$.

\subsection{\ourmodel}
We propose a novel diffusion transformer adapter that seamlessly integrates speaker speech and motion in a temporally causal manner. At the core of \ourmodel are the CTM-Adapter layers, with our model structured as a hierarchical stack of these modules to effectively capture multimodal dependencies and generate coherent listener responses.\\
\noindent \textbf{CTM-Adapter Blocks.} Inspired by IP-Adapter~\cite{ye2023ip} for UNet-based diffusion models~\cite{rombach2022high}, we propose CTM adapter -- a video diffusion adapter that extends adapters from conventional spatial attention to multimodal temporal attention for better temporally-aligned guidance in a Multimodal Video DiT. Built upon the MM-DiT Full Attention blocks from EasyAnimate \cite{xu2024easyanimate}, our CTM-Adapter blocks introduce two additional Temporal Causal Attention (TCA) modules to process temporally fused speech and motion features. The outputs of the standard MM-DiT Full Attention module are then combined with the speech-guided and motion-guided cross-attention features produced by our TCA modules using a weighted summation, effectively capturing the temporal dependencies between multimodal inputs. Following L2L \cite{ng2022learning}, we apply causal masks to our TCA modules so that current listener motions are only generated with respect to past speaker motions and audio. 

Formally, the original MM-DiT Full Attention block takes visual and textual tokens as input, denoted as $X_v$ and $X_t$, respectively. In our case, $X_v$ represents the generated listener video frames. For appearance control, the first CTM-Adapter block is modified to accept concatenated inputs $[X_v, X_i]$, where $X_v$ is initialized with random noise. 
The text-video tokens are then concatenated and passed through a multi-layer perceptron (MLP), after which they are split back into text tokens and denoised video tokens. These tokens are subsequently fed into the next layer for further processing. The attention mechanism within the MM-DiT blocks is given by
\begin{align}
  \text{Attn}_{\text{MMDiT}}\left([X_tX_v]\right) 
  &= \sigma\left( 
      \frac{\left[Q_tQ_v\right][K_tK_v]^\top}{\sqrt{d}} 
  \right) \nonumber \\
  &\quad \times [V_tV_v]
  \label{eq1}
  \end{align}
\noindent
where $\sigma$ is a Softmax function, $[Q_tQ_v],[K_tK_v],[V_tV_v]$ are the concatenation of query, key, and value projections from text $X_t$ and video $X_v$ tokens respectively, and $d$ is the projection dimension.

To adapt MM-DiT for the task of listener behavior generation with additional speech and motion inputs, we leverage the Decoupled Multimodal Attention mechanism \cite{esser2024scaling} to integrate speech-guided and motion-guided information into the attention modules of MM-DiT. Specifically, 
\begin{align}
  \text{Attn}_{\text{speech}}\left([X_tX_v], X_s\right) 
  = \;& \sigma\Bigg(\frac{ M \circ \left[Q_tQ_v\right] K_s^\top }{\sqrt{d}} \Bigg) V_s
  \end{align} 
\begin{align}
  \text{Attn}_{\text{motion}}\left([X_tX_v], X_m\right) 
  = \;& \sigma\Bigg(\frac{ M \circ \left[Q_tQ_v\right] K_m^\top }{\sqrt{d}} \Bigg) V_m
  \end{align}
where $[K_s, K_m]$ and $[V_s, V_m]$ are the key and value projections of $X_s$ and $X_m$ respectively, and $M$ correspond to the causal attention masks. Attention is computed along the temporal dimension, while the spatial dimension is combined with the batch dimension. The outputs of the three attention branches are combined via an element-wise summation
\begin{align}
   \text{Attn}_{\text{CTM-adapter}}\left([X_tX_v]\right)  = \;& \alpha \times \text{Attn}_{\text{speech}}\left([X_tX_v], X_s\right) \nonumber \\
  &+\; \beta \times \text{Attn}_{\text{MMDiT}}\left([X_tX_v]\right) \nonumber \\
  &+\; \gamma \times \text{Attn}_{\text{motion}}\left([X_tX_v], X_m\right)
  \end{align}
where $\alpha, \beta, \gamma$ are scaling parameters that control the contribution of each branch during inference. 

Overall, \ourmodel-Gen consists of a stack of $L$ CTM-Adapter Blocks, trained using the standard diffusion loss similar to  \cite{xu2024easyanimate}.
\\
\noindent \textbf{Causal Long Sequence Generation.} \ourmodel-Gen is designed to generate segments of \( K \) frames at a time. When synthesizing listener behavior for input sequences longer than \( K \) frames, we partition the inputs into overlapping segments of \( K \) frames and generate each segment independently. However, directly merging these segments often results in abrupt discontinuities in pose and expression due to the lack of temporal consistency across segment boundaries. To address this, we introduce \textit{\ourmodel-Edit} to produce smooth transitions between consecutive segments.  

\ourmodel-Edit operates by selecting two consecutive segments of length \( K \) generated by \ourmodel and extracting the last \( K' \) (\( K' \ll K \)) frames from the first segment and the first \( K' \) frames from the second segment. While these segments should ideally form a seamless motion sequence, independent generation can introduce unnatural transitions. To refine these transitions, we pass the first and the last of concatenated \( 2K' \) frames through a standard DiT model, with the objective of reconstructing the corresponding ground-truth frames for these \( 2K' \) timesteps. The model is trained as a standard diffusion model, learning to synthesize temporally smooth and coherent frames that effectively bridge discontinuities between independently generated segments. 

\noindent \textbf{Customized text control.} Existing dyadic behavior learning datasets lack annotated descriptions of listener behavior, making them unsuitable for text-guided listener behavior generation. To bridge this gap, we leverage Google Gemini 1.5 \cite{gemini} to extract affective-related descriptions of listener behavior from existing datasets. These generated textual descriptions are then used to train \ourmodel, enabling it to learn text-aware behavior generation. During inference, users can provide custom textual descriptions to guide the model in producing personalized listener responses. Details of our text prompts are provided in the supplementary materials.

\begin{figure*}
    \centering
    \includegraphics[width=\linewidth]{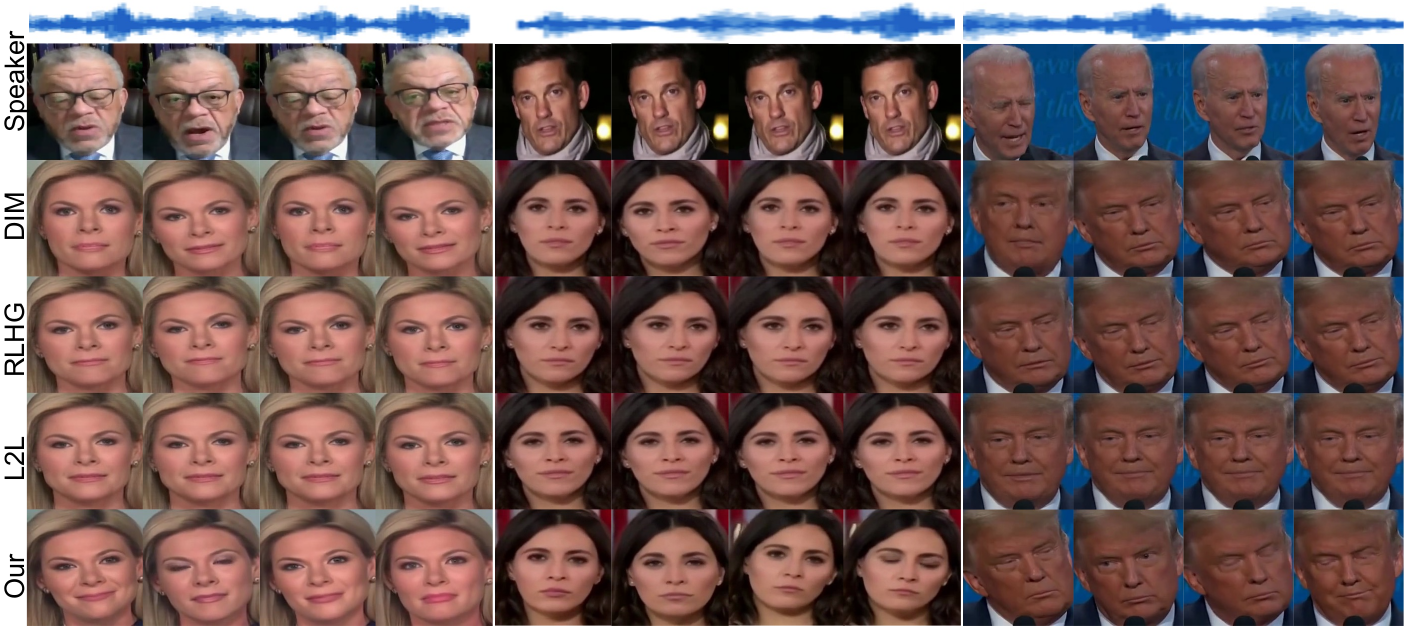} 
    \caption{\textbf{Qualitative Comparison on ViCo test set.} Our method generates high-quality, photorealistic facial images with diverse and natural social behaviors, including head movements and blinks, whereas baseline methods often produce less varied and expressive responses. }
    \label{fig:comp_vico} 
    \vspace{-1em}
  \end{figure*}

  \begin{figure}
    \centering
    \includegraphics[width=\linewidth]{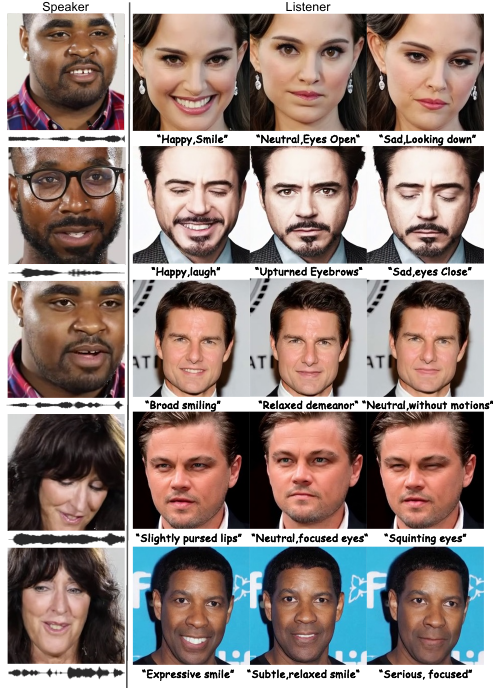} 
    \caption{\textbf{Listener generation from \ourmodel on out-of-domain identities.} Our method can integrate expressions from text conditions and synthesize diverse responses to the speakers.}
    \label{fig:vis_ood} 
  \end{figure}

%% file: sec/4_exp.tex
 
\begin{table}[h]
\centering

\scalebox{0.85}
{\begin{tabular}{lcccccccccccc}
\toprule
Method  &  Feedback$\uparrow$ &  Diversity$\uparrow$ &  Smoothness$\uparrow$ & Overall$\uparrow$ \\
\midrule
ELP~\cite{song2023emotional}  & 2.93\% & 4.04\% & 5.65\% & 4.21\% \\
RLHG~\cite{zhou2022responsive}  & \underline{10.45\%} & \underline{12.14\%} & 13.00\% & 11.86\%\\
 L2L~\cite{learning2listen}  & 4.20\% &6.06\% & 10.95\% & 7.07\%\\
  DIM~\cite{tran2024dim}  & 10.25\% & 12.12\% &\underline{16.11\%} & \underline{12.83\%} \\
Ours  &\textbf{72.17\%} & \textbf{65.64\%} &\textbf{54.29\%} & \textbf{64.03\%} \\
\bottomrule
\end{tabular}}

\caption{User Study. We ask the participants to choose the best video among all methods for each criterion. The average preference percentage for each method output is provided.}
\label{tab:user_study}
\end{table}


  \begin{table}[h!]
    \centering
        \scalebox{0.85}{
    \begin{tabular}{@{}lccccc@{}}
      \toprule
      \textbf{Method} & \textbf{FID$\downarrow$} & \textbf{FVD$\downarrow$} & \textbf{LPIPS$\downarrow$} & \textbf{PSNR$\uparrow$} & \textbf{SSIM$\uparrow$} \\
      \midrule

      RLFG~\cite{learning2listen} & 52.01 & 287.48 & 0.37 & 17.95 & 0.61 \\
      L2L~\cite{learning2listen} & 56.93 & 285.24 & 0.40 & 18.01 & 0.64 \\
      ELP~\cite{song2023emotional} & 67.12 & 281.95 & 0.38 & 17.96 & 0.61 \\
        DIM~\cite{tran2024dim}  & 49.75 & 288.27 & \underline{0.36} & 18.51 & 0.62 \\
            \midrule
      w/o Text-Control& \underline{10.56} & 57.71 & \textbf{0.28} & \underline{20.58} & \underline{0.71} \\
      w/o CTM-Adapter& 11.56 & \underline{54.79 }& \textbf{0.28} & \textbf{20.77} & \textbf{0.72} \\
      \ourmodel & \textbf{10.14} & \textbf{53.54} & \textbf{0.28} & \textbf{20.77} & \underline{0.71} \\
      \bottomrule
    \end{tabular}}
    \caption{Quantitative comparison on VICO~\cite{yu2023responsive} test set. The metrics are reported in photorealistic video space.}
    \label{tab:visual_comparison_vico}
  \end{table}
  
  \begin{table}[h!]
    \centering
    \scalebox{0.85}
{    \begin{tabular}{@{}lccccc@{}}
      \toprule
      \textbf{Method} & \textbf{FID$\downarrow$} & \textbf{FVD$\downarrow$} & \textbf{LPIPS$\downarrow$} & \textbf{PSNR$\uparrow$} & \textbf{SSIM$\uparrow$} \\
      \midrule
      RLHG~\cite{yu2023responsive} & 30.55 & 217.48 & 0.41 & 17.23 & 0.53 \\
      L2L~\cite{learning2listen} & 56.43 & 522.67 & 0.46 & 16.49 & 0.53 \\
      DIM~\cite{tran2024dim} & 30.75 & 204.49 &0.40 &17.74 &0.54 \\ 
      \midrule
      
      w/o Text-Control & \underline{8.05} & 49.88 & \underline{0.27} & \underline{18.70}
      & \underline{0.59} \\
      w/o CTM-Adapter & 8.16 & \underline{45.62} & \underline{0.27}&  18.60 & 0.55 \\
      \ourmodel & \textbf{7.99} & \textbf{45.42} & \textbf{0.26} & \textbf{18.73} &  \textbf{0.60} \\ 

      \bottomrule
    \end{tabular}}
    \caption{Quantitative comparison on RealTalk~\cite{geng2023affective} test set. The metrics are reported in photorealistic video space.}
    \label{tab:visual_comparison_realtalk}
  \end{table} 

\section{Experiments}\label{sec:experiments}
\subsection{Implementation Details}
\par \noindent \textbf{Datasets}
We evaluate \ourmodel on RealTalk \cite{geng2023affective} and ViCo \cite{vicox} datasets. RealTalk \cite{geng2023affective} is a large dataset containing 692 dyadic conversations with a total duration of $115$ hours. To allow comparison with existing methods, we also evaluate our method on ViCo \cite{zhou2022responsive}, which is a smaller dataset consisting of 483 video sequences featuring 76 unique listeners. We present the details of dataset pre-processing in the supplementary material.

\par \noindent \textbf{Model Training and Inference} We used a pre-trained DiT-based video diffusion model as the backbone, with initialization weights from EasyAnimateV5.1-12b~\cite{xu2024easyanimate}. Causal Motion and Speech Encoders are transformers trained from scratch sharing the same architecture with 6 attention blocks, 8 heads, 512 hidden dimensions. For CTM-Adapter we set $\gamma$, $\beta$, and $\alpha$ parameters to be 1.0.  We trained our model on RealTalk~\cite{geng2023affective} with 8 NVIDIA H100 GPUs for 20k iterations with a batch size of 8. The training videos were divided into 4-second clips consisting of 144 frames, sampled at a stride of 3 frames. Each frame was resized to $256 \times 256$ pixels. We used the Adam optimizer with a learning rate of $2 \times 10^{-5}$, a weight decay of $5 \times 10^{-3}$, and gradient clipping of 0.05. The entire model was trained end-to-end, with text control sequences truncated at 512 tokens per video. Text is encoded and passed to the model via T5~\cite{2020t5} and BERT~\cite{devlin2019bert} encoder. For DiTaiListener-Edit we set \( K \) to 49, \( K' \) to 6. To compare with baselines on ViCo~\cite{yu2023responsive}, we train our \ourmodel for 5k iterations with the same hyperparameters. 

\subsection{Evaluations and Comparisons}

\par\noindent\textbf{Baselines} We tested four listener behavior generation methods to benchmark our model against, including:
\begin{itemize}
    \item \textbf{RLHG~\cite{zhou2022responsive}} is an RNN-based listener prediction model.
    \item \textbf{L2L~\cite{learning2listen}} uses quantized VQ-VAE to predict listener motions in an autoregressive manner.
    \item \textbf{ELP~\cite{song2023emotional}} is a model that relies on emotion vectors and speaker features to predict listener behavior.
    \item \textbf{DIM~\cite{tran2024dim}} is a recent method that combines two-branch speaker-listener VQ-VAE with contrastive pretraining.
    \item \textbf{INFP~\cite{zhu2024infp}} is a recent diffusion-based model which learns a motion latent space trained on in-house data. Currently, the code and data are not open-sourced.
\end{itemize}
 Note that these methods are all multi-stage models, which predict discrete listener motion representation first, then map these predictions into photorealistic videos via PIRenderer~\cite{ren2021pirenderer} or other similar rendering methods \cite{Drobyshev2022}.

 \begin{table*}[ht!]
    \centering
    
    \resizebox{0.65\textwidth}{!}{
    {\begin{tabular}{lcccccccccc}

    \toprule
    \multirow{2}[2]{*}{Method}  & \multicolumn{2}{c}{\textbf{FD}$\downarrow$} & \multicolumn{2}{c}{\textbf{P-FD}$\downarrow$} & \multicolumn{2}{c}{\textbf{MSE}$\downarrow$} & \multicolumn{2}{c}{\textbf{SID}$\uparrow$} & \multicolumn{2}{c}{\textbf{Var}$\uparrow$}  \\ 
    \cmidrule(lr){2-3} \cmidrule(lr){4-5}  \cmidrule(lr){6-7}  \cmidrule(lr){8-9}  \cmidrule(lr){10-11} 
     &  \textbf{Exp}   & \textbf{Pose} & \textbf{Exp}   & \textbf{Pose}   & \textbf{Exp}   & \textbf{Pose}  & \textbf{Exp}   & \textbf{Pose} & \textbf{Exp}   & \textbf{Pose} \\
    \midrule
     ELP~\cite{song2023emotional}&47.17 & 0.08 & 47.48 & 0.08&0.98& \underline{0.02}& 1.76&1.66& 1.49&\underline{0.02} \\
     RLHG~\cite{zhou2022responsive} & 39.02&0.07 &40.18 &0.07  &0.86 &\textbf{0.01} &3.62 &3.17 &1.52 &\underline{0.02} \\
    
    L2L~\cite{ng2022learning} & 33.93 & 0.06 &35.88 & \underline{0.06}&0.93 &\textbf{0.01} &2.77 &2.66 &0.83 &\underline{0.02} \\
    
    DIM~\cite{tran2024dim} &23.88 &0.06 & 24.39 & \underline{0.06} & \underline{0.70} &\textbf{0.01} &3.71 &2.35 &1.53 &\underline{0.02} \\
    
    INFP$\dagger$~\cite{zhu2024infp} &\underline{18.63} & 0.07 & - & - & \textbf{0.51} & \textbf{0.01} & \underline{4.78} & \textbf{3.92} &\textbf{2.83} &\textbf{0.18} \\
    
    GT & - & -& - & -& - & - &5.03 &4.07 &0.93 &0.01 \\
    \midrule
    w/o Text-Control & 18.88 & \underline{0.05} & \underline{21.64} & \underline{0.06} & 0.85 & \underline{0.02} & \textbf{4.81} & 3.59 & 1.43 & \underline{0.02}\\
    w/o CTM-Adapter & 19.54 & \underline{0.05} & 22.27 & 0.06 &  0.88 & \underline{0.02} & 4.68 & 3.71 & 1.46 & \underline{0.02} \\
    \ourmodel & \textbf{17.49} & \textbf{0.04} & \textbf{20.53} & \textbf{0.05} & 0.85 & \underline{0.02} & 4.73 & \underline{3.74} & \underline{1.51} &  \underline{0.02} 
   \\ 
    \bottomrule
    \end{tabular}}}
    \vspace{-5pt}
    \caption{Quantitative comparison on ViCo~\cite{zhou2022responsive} test set in the 3DMMs (EMOCA~\cite{emoca}) space. $\dagger$ denotes the method did not release code or in-house training data, and the numbers are directly taken from their paper.}
    \label{tab:quant_comp_vico}
    \end{table*}

  \begin{table*}[t!]
    \centering
    \resizebox{0.65\textwidth}{!}{
    {\begin{tabular}{lcccccccccc}

    \toprule
    \multirow{2}[2]{*}{Method}  & \multicolumn{2}{c}{\textbf{FD}$\downarrow$} & \multicolumn{2}{c}{\textbf{P-FD}$\downarrow$} & \multicolumn{2}{c}{\textbf{MSE}$\downarrow$} & \multicolumn{2}{c}{\textbf{SID}$\uparrow$} & \multicolumn{2}{c}{\textbf{Var}$\uparrow$}  \\ 
    
    \cmidrule(lr){2-3} \cmidrule(lr){4-5}  \cmidrule(lr){6-7}  \cmidrule(lr){8-9}  \cmidrule(lr){10-11}  
     &  \textbf{Exp}   & \textbf{Pose} & \textbf{Exp}   & \textbf{Pose}   & \textbf{Exp}   & \textbf{Pose}  & \textbf{Exp}   & \textbf{Pose} & \textbf{Exp}   & \textbf{Pose} \\
    \midrule
     RLHG~\cite{zhou2022responsive} & 69.04 & \underline{0.05} & 69.09 & \underline{0.06}  & 1.37 & \textbf{0.01} & 0.35 & 3.23 & 0.14 & \textbf{0.01 }
     \\
    
    L2L~\cite{ng2022learning} & 72.89 & 0.10 & 72.94 & 0.10 & 1.44 & \underline{0.02} & 0.10 & 2.42 & 0.07 & \textbf{0.01} \\
    DIM~\cite{tran2024dim} & 77.97 & 0.15 & 78.70 &  0.15 & 1.52 &  \underline{0.02} & 3.49 & 3.29 & 0.74 &  \textbf{0.01} \\
    GT & - & - & - & - & - & - & 5.13 & 3.95 & 1.36 & 0.02 \\
    \midrule
    
    w/o Text-Control& 15.62 & \textbf{0.02} & 16.35 & \textbf{0.02} & 0.67 & \textbf{0.01} & 5.01 & 3.94 & \underline{1.30} &  \textbf{0.01} \\
    w/o CTM-Adapter & \underline{15.10} & \textbf{0.02} & \underline{15.82} & \textbf{0.02} & \underline{0.66} & \textbf{0.01} & \textbf{5.11} & \textbf{3.98} & \underline{1.30} & \textbf{0.01} \\
    \ourmodel & \textbf{14.28} & \textbf{0.02} & \textbf{15.07} & \textbf{0.02} & \textbf{0.65} & \textbf{0.01} & \underline{5.09} & \underline{3.95} & \textbf{1.31} & \textbf{0.01} \\
    \bottomrule
    \end{tabular}}}
    \vspace{-5pt}
    \caption{Quantitative comparison on RealTalk~\cite{geng2023affective} test set in the 3DMMs (EMOCA~\cite{emoca}) space.}
    \label{tab:quant_comp_realtalk}
    \vspace{-1em}   
    \end{table*}
    
\par \noindent 
 \textbf{Evaluation metrics.}
Since our method is an end-to-end model that generates photorealistic videos directly, we first evaluate its performance against baseline methods using image-space metrics: Peak Signal-to-Noise Ratio (PSNR), Learned Perceptual Image Patch Similarity (LPIPS)~\cite{zhang2018perceptual}, SSIM~\cite{wang2004image}, and Fréchet Inception Distance (FID). PSNR~\cite{hore2010image} is a pixel-wise image similarity metric. LPIPS and SSIM are perceptual image-based measures that evaluate the similarity between predictions and ground-truth frames. Fréchet Inception Distance~\cite{heusel2017gans} is used to measure how different the distributions of the generated frames are from real data. Fréchet Video Distance~\cite{unterthiner2019fvd} compares distributions of generated videos with the distribution of ground-truth videos, including temporal information. Following previous methods~\cite{learning2listen,tran2024dim}, we also compare the results in the 3DMM space by extracting EMOCA~\cite{emoca} parameters from our generated frames. We use Frechet Distance (FD), Mean Squared Error (MSE), Paired FD (P-FD), Variance (Var), and SI Diversity (SID) as evaluation metrics. Details of these metrics are available in the supplementary material. Note that Var and SID are metrics to evaluate the diversity of data distribution, so we are able to report these for Ground Truth EMOCA parameters as well. FD and P-FD are the primary metrics to evaluate the listener generation quality. 

\par \noindent 
 \textbf{Quantitative Comparison}\label{quantitative}
We compare our method to the state-of-the-art listener video generation baselines~\cite{learning2listen,zhou2022responsive,song2023emotional,tran2024dim,zhu2024infp}. This work focuses on improving the quality of motions and the photorealism of the generated video.    
Table~\ref{tab:visual_comparison_vico} and Table~\ref{tab:visual_comparison_realtalk} present the quantitative analysis of such quality on ViCo and RealTalk datasets. \ourmodel achieves significant improvements across different baseline models, indicating that the proposed method generates vivid expressiveness of listener head videos. The baseline methods heavily rely on the renderer and are not able to directly generate high-fidelity human portraits.
Since previous work~\cite{learning2listen,tran2024dim,song2023emotional,zhou2022responsive} widely adopted 3DMM~\cite{DECA:Siggraph2021,emoca} to represent pose and facial expression and provided evaluation in such parameter space, we extract EMOCA~\cite{emoca} parameters from our generated photorealistic videos and presents a quantitative analysis in
Tables~\ref{tab:quant_comp_vico} and~\ref{tab:quant_comp_realtalk}. Even though our model is not trained to generate 3DMM parameters, it achieves competitive performance with the existing methods, which indicates that \ourmodel generates high-quality videos with diverse listener motions that provide correct feedback to the speakers.

 \par \noindent 
 \textbf{Qualitative Comparison}\label{qualitative}
We qualitatively compare the generated listener video from \ourmodel with previous methods~\cite{learning2listen,tran2024dim,song2023emotional,zhou2022responsive} on ViCo in Figure~\ref{fig:comp_vico}. Note that these works used 3DMMs as predictions from their models, we render the photorealistic results by PIRenderer~\cite{ren2021pirenderer} following the setting mentioned in~\cite{learning2listen,tran2024dim}. All baselines exhibit limited expressiveness, with most of their generated listener appearing almost static. Please refer to additional video examples provided in the supplementary materials for clearer observations and further comparison.
We also provide more visualizations of text-control from \ourmodel on out-of-domain data in Figure~\ref{fig:vis_ood}. Our method demonstrates user-friendly customization of listener emotions through natural language and strong generalizability to other unseen identities. We provide more video visualizations in the supplementary offline page.
  
\par \noindent 
 \textbf{User Study}
We asked 50 participants from \textbf{Prolific}, a crowdsourcing platform, to evaluate the quality of the listeners generated with the ViCo test set~\cite{zhou2022responsive} using L2L~\cite{ng2022learning}, RLHG~\cite{zhou2022responsive}, ELP~\cite{song2023emotional}, DIM~\cite{tran2024dim} and \ourmodel. The methods were anonymized and randomly ordered for each question. The average duration of the videos is $14$ seconds. For each group of video comparisons, we ask the user to choose the best method according to the following criteria for judgment: 1) The response from the listener provides correct feedback to the speaker. 2) The head motions and facial expressions are diverse. 3) The general video quality is flawless and smooth. The user study demonstrated that most users selected our model output as superior across all scales (see Table~\ref{tab:user_study}). We present an example from our user study survey in Figure.~\ref{fig:user_study} of the supplementary material.\\

 \subsection{Ablation Analysis}

\begin{figure}[ht]
    \centering
    \includegraphics[width=\linewidth]{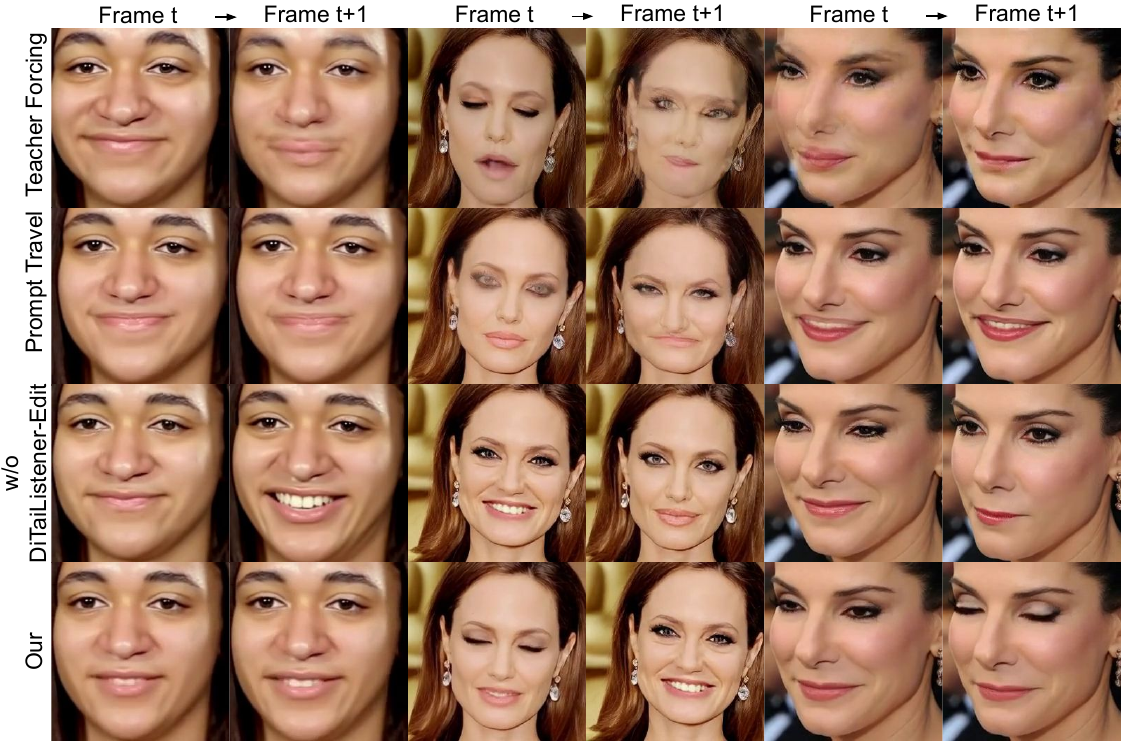} 
    \vspace{-15pt}
    \caption{\textbf{Qualitative comparison of long video generation.} Our method generates smoother videos with fewer transition artifacts compared to prompt traveling and teacher forcing methods.}
    \label{fig:long_video_comp} 
  \end{figure}

In this section, a comprehensive ablation analysis of \ourmodel is presented. We evaluate the effectiveness of our proposed Causal Multimodal Temporal Adapter and text control on ViCo and RealTalk test sets in Tables~\ref{tab:visual_comparison_vico},~\ref{tab:visual_comparison_realtalk},~\ref{tab:quant_comp_vico} and~\ref{tab:quant_comp_realtalk}. Note that in all these tables, each row represents the individual ablation. \textbf{w/o Text-Control} means an empty string replaces the input to text encoders, and \textbf{w/o CTM-Adapter} denotes the Causal Temporal Attention is replaced by simple non-causal spatial attention without masks.
To further confirm the effectiveness of text control, we show videos of different listener emotions on the supplementary offline page. We also visualize the comparison between our proposed Causal Long Sequence Generation strategy, frame-smoothing \textbf{\ourmodel-Edit}, with widely used prompt travel and teacher forcing~\cite{williams1989learning,guo2023animatediff} in Figure.~\ref{fig:long_video_comp} and the supplementary offline page. Note that \textbf{w/o \ourmodel-Edit} in both figures and offline webpage denotes the video segments are directly concatenated in time dimension without smooth processing with \ourmodel-Edit.
\textbf{Teacher Forcing}~\cite{williams1989learning} produces longer videos recurrently relying on its own predictions by replacing the appearance frame with the last generated frame. Teacher forcing suffers from error accumulation and gradual information loss, leading to quality degradation for long sequences. \textbf{Prompt Travel} relies on a sliding window approach for long video generation. The method starts by generating long noise patterns. During each denoising step, a sliding window is taken over overlapping chunks of video; denoising predictions are made for each chunk independently. Predictions from overlapping chunks are averaged and passed to the next denoising step. Prompt Travel requires additional resources proportional to overlap length. While effective against error accumulation, it suffers from blurring artifacts due to averaging and transition artifacts, as the model is not trained on inputs produced via averaging.

%% file: sec/5_conclusion.tex
\section{Conclusion}
We introduce \ourmodel, a DiT-based framework for generating high-fidelity listener response videos from speaker audio and facial motion inputs. Unlike prior methods that rely on intermediate 3DMM representations, \ourmodel\ directly synthesizes photorealistic listener portraits in an end-to-end manner while enabling text-controlled customization of listener behaviors. At its core, the Causal Temporal Multimodal Adapter (CTM-Adapter) effectively integrates multimodal inputs in a temporally causal manner. To ensure smooth transitions between generated video segments, we proposed \textit{\ourmodel-Edit}, which refines segment boundaries for continuous and natural listener behaviors over long sequences. Our model demonstrates state-of-the-art performance in producing expressive, semantically aligned listener responses with high visual fidelity and temporal consistency. This work advances AI-driven human interaction modeling and has broad applications in virtual avatars, human-computer interaction, and social robotics. Future directions include expanding listener behavior diversity, improving real-time inference, and integrating more contextual cues for enhanced responsiveness.\\
\noindent\textbf{Limitations and Future Works}
Despite its effectiveness in the expressiveness generation of listener videos, our \ model has certain limitations, particularly in inference efficiency. In the future, we will explore further techniques to accelerate the sampling procedure and provide faster video generation.\\
\noindent\textbf{Ethics Statement.} Our work aims to improve human behavior generation from a technical perspective and is not intended for malicious use like impersonation. Any future application should implement safeguards to ensure consent from the people whose likeness is being generated, and synthesized videos should clearly indicate their artificial nature.

%% file: sec/ackn.tex
\section*{Acknowledgments}
Research was sponsored by the Army Research Office and was accomplished under Cooperative Agreement Number W911NF-25-2-0040. The views and conclusions contained in this document are those of the authors and should not be interpreted as representing the official policies, either expressed or implied, of the Army Research Office or the U.S. Government. The U.S. Government is authorized to reproduce and distribute reprints for Government purposes notwithstanding any copyright notation herein

%% file: sec/X_suppl.tex
\clearpage
\setcounter{page}{1}
\maketitlesupplementary

\section{Detailed User Study}
\begin{figure}

    \centering
    \includegraphics[width=\linewidth]{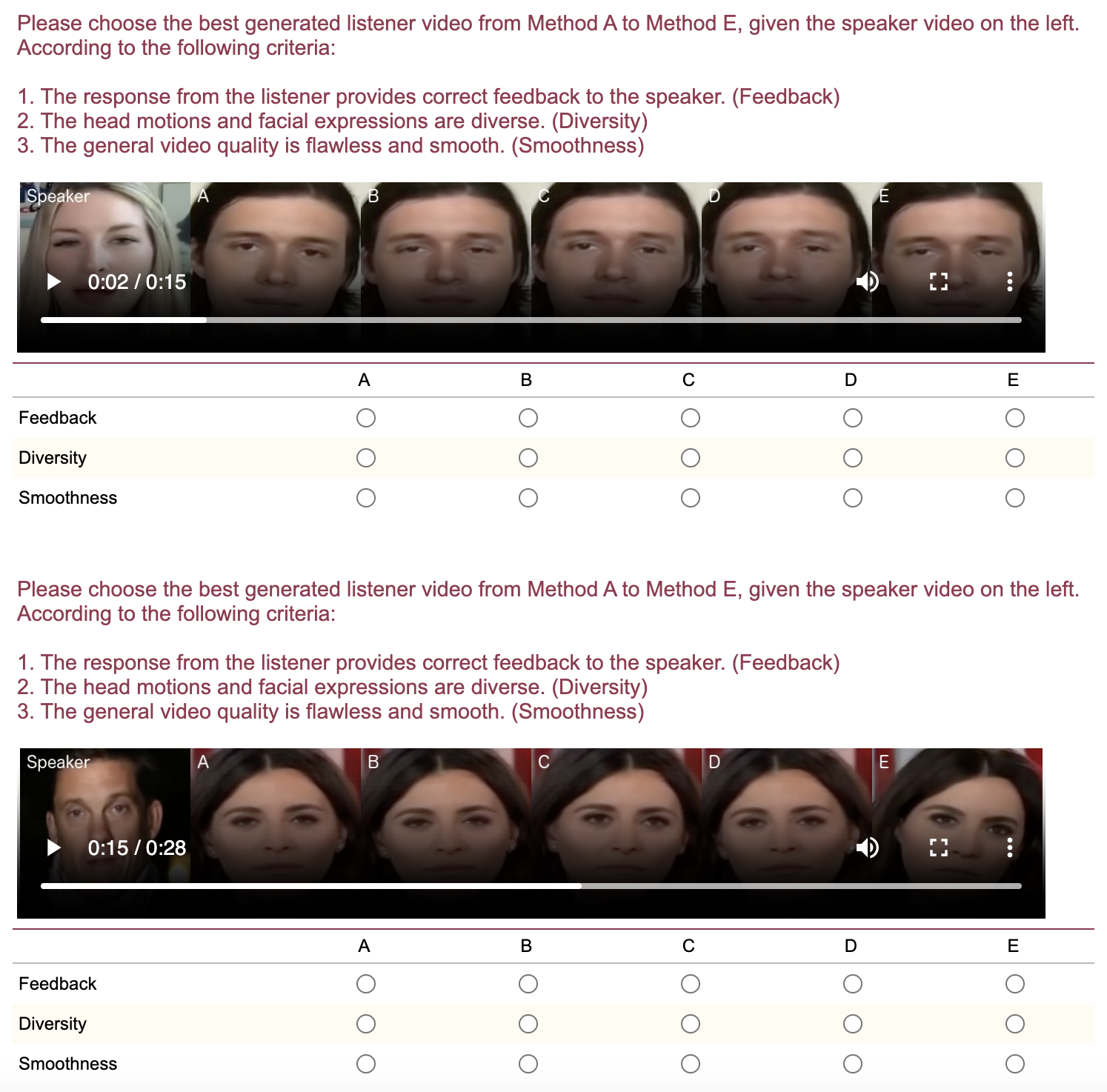} 
    \caption{A screenshot of user study survey example. The methods are anonymized as A, B, C, D, E, and the order is randomized.}
    \label{fig:user_study} 
  \end{figure}

We show an example screenshot of our user study survey in Figure.~\ref{fig:user_study}. The methods are well-anonymized, and we keep track of the original method order for each comparison offline. The participants are paid \$12/hr for their labor, and we collect the results of each user after careful human review. The incomplete and invalid responses are rejected. After such a review, we have 50 valid responses in total. The result is presented in Table.~\ref{tab:user_study}



\section{Details of Dataset Processing}
\textbf{RealTalk} database includes videos from The Skin Deep YouTube Podcast and contains 692 dyadic conversations. The total duration of the videos is $115$ hours. The database provides bounding boxes of participants' faces, as well as EMOCA 3DMM coefficients \cite{emoca}. The dataset contains around 50,000 conversational turns of dyadic interactions with specified roles, i.e., speaker vs listener. We noticed that some of the videos contain artifacts that make them hard to use for direct training in the video-generation model; such artifacts include occlusions between the listener and the camera, bounding boxes not covering full faces, shaky bounding boxes, bounding boxes changing size, and extreme head poses. 

Training the model on raw data led to model collapse (floating hands, blurry, distorted faces, face out of frame), necessitating data cleaning. To address this, we applied a multi-step filtering process. First, for each clip, instead of using per-frame bounding boxes, we identified the largest bounding box covering faces in all frames and used it for cropping. This provided a static background and a constant bounding box size and removed frame-to-frame jitter. We then used LibreFace~\cite{chang2024libreface} to estimate head pose and only kept frames with frontal-looking faces by applying pitch and yaw thresholds; the acceptable range of yaw and pitch angles was set to $[-30, 30]$ degrees. 
In order to remove other frame-level artifacts, we collected annotations in-house. For every one of the 50,000 clips, we randomly selected a frame. Annotators were to label if each frame had artifacts of some category: split screens, text overlays or transition effects, non-hand objects blocking the view, visible hands, and other visual disturbances such as glare. Frames without artifacts were automatically labeled as clean. 
Initial manual sparse annotations were used to finetune a CLIP image encoder for artifact detection. This method provided 0.95 recall on unseen clips. As a result, 13.6\% of all data was marked for removal due to detected artifacts. The resulting model was applied with a high sensitivity threshold to filter artifacts on all the other frames in the dataset.   


\section{Details of 3DMM Space Evaluation Metrics}

\begin{itemize}

\item Fréchet Distance (FD) measures discrepancies between distributions of real and generated listener's EMOCA 3DMM coefficients. It is separately computed for head poses and facial expressions.
\item Paired FD (P-FD) is an extension of FD for synchrony. P-FD is calculated by concatenating listener motions with GT speaker motions and measuring Fréchet Distance of resulting features. It measures how well listener motions are temporally aligned with speaker motions.
\item Variance (Var) and SI for Diversity (SID) are used to measure the diversity of generated listener motions, following DIM~\cite{tran2024dim} and L2L \cite{learning2listen}. SID is computed by applying k-means clustering to 3DMM features and computing entropy of the histogram of cluster assignments. The higher the SID, the more diverse the motions are. Var is the variance of pose and expression features that indicate how expressive the movements are across temporal dimensions. 

 \section{Text control} 
 We add text control inputs to guide the model with text prompts. ViCo and RealTalk datasets do not contain any textual captions. Therefore, we performed captioning in two stages: text extraction and prompt refinement. In the first stage, we used a Large Video-Language-Model (VLM) Google Gemini 1.5 Flash 002 \cite{gemini} to extract text descriptions of audio-visual emotional cue information from the listener video with the speaker audio. The following prompt was used: 
 \begin{alltt}
Which emotion is present in the video?
Emotion can belong to only one of the
following classes - angry, happy, sad,
neutral, disgust, fear, surprise. Give
your response in JSON format as 
following: \{"emotion": "angry", 
"reason":"reason for your response"\}
\end{alltt} 
 Example of model respose:
 \begin{alltt}
\{"emotion": "happy", "reason": "The 
person in the video is expressing 
feelings of freedom and self-
acceptance. The overall tone is
positive, suggesting contentment 
and happiness. He talks about being
able to be himself without 
conforming to external pressures. 
This indicates a sense of 
liberation and joy. Phrases such as 
"I'm free" and "I get to be me" 
directly convey positive emotions."\}
 \end{alltt}
 Extracted information has emotional cues from the speech and quotes that are important to describe the context of the conversation. However, those do not directly affect the appearance of the listener; sometimes, the VLM lists what emotions are not present in the video. Therefore, an extra step of text processing was needed to create a text control prompt from extracted information. We use Llama-3.3-70B-Instruct Large Language Model (LLM) \cite{touvron2023llama} with the following prompt:
 \begin{alltt}
 """
Instructions:
Given an analytical description of a 
dyadic conversation with a video of a
listener, extract only information 
about the listener's physical 
expressions and emotions, conversation
emotions. Discard conversation quotes, 
reasoning thoughts, and any general 
explanations or definitions of 
emotions. Only retain direct 
observations about the listener's 
facial expressions, emotions, or other 
reactions. Give only extracted 
information in plaintext form in a 
single sentence.

Analytical description example:
"The person in the video is shown 
smiling broadly, with their mouth wide
open and eyes crinkled at the corners.
This is a classic display of happiness
There are no other indicators of any 
other emotion present."

Example extracted: Listener looks 
happy, smiling broadly with mouth open 
and eyes crinkled.

User:
Analytical description:
"\{text\}"

Assistant:
Extracted: Listener looks """
 \end{alltt}
 The processed text is used as text control. Example of resulting text control: ``happy, smiling broadly, showing her teeth".

\end{itemize}
